# The Powerful Use of AI in the Energy Sector: Intelligent Forecasting

Erik Blasch[1], Haoran Li[2], Zhihao Ma[2], Yang Weng[2]

[1]MOVEJ Analytics, OH, erik.blasch@gmail.com
[2] Arizona State University, AZ, {lhaoran,zhihaoma,yang.weng}@asu.edu

## Abstract

Artificial Intelligence (AI) techniques continue to broaden across governmental and public sectors, such as power and energy - which serve as critical infrastructures for most societal operations. However, due to the requirements of reliability, accountability, and explainability, it is risky to directly apply AI-based methods to power systems because society cannot afford cascading failures and large-scale blackouts, which easily cost billions of dollars. To meet society requirements, this paper proposes a methodology to develop, deploy, and evaluate AI systems in the energy sector by: (1) understanding the power system measurements with physics, (2) designing AI algorithms to forecast the need, (3) developing robust and accountable AI methods, and (4) creating reliable measures to evaluate the performance of the AI model. The goal is to provide a high level of confidence to energy utility users. For illustration purposes, the paper uses *power system event forecasting* (PEF) as an example, which carefully analyzes synchrophasor patterns measured by the Phasor Measurement Units (PMUs). Such a physical understanding leads to a data-driven framework that reduces the dimensionality with physics and forecasts the event with high credibility. Specifically, for dimensionality reduction, machine learning arranges physical information from different dimensions, resulting inefficient information extraction. For event forecasting, the supervised learning model fuses the results of different models to increase the confidence. Finally, comprehensive experiments demonstrate the high accuracy, efficiency, and reliability as compared to other state-of-the-art machine learning methods.

## 1. Introduction

Uncertain renewable resources and loads increasingly challenge government and public electricity supports. The uncertainties cause difficulties for energy management and stability guarantees. Unlike many conventional generation sources, many renewable resources in public sectors, e.g., wind power and photovoltaic (PV) solar power, are considered variable generation (VG). They have a maximum generation limit that changes with time and the limit is uncertain. Variability of VG occurs at multiple timescales, which may cause an unbalance of power generation and load [1]. Due to these issues, events may often happen in the public grids. For example, intermittent renewable energies can create both two-way power flow and system oscillations [2]. Large penetrations of electric vehicles (EVs) in a short timeframe may also cause energy line trips due to a long-time overload [3].

To handle issues of stable power and energy availability, it's important to increase the situational awareness of the public energy system through advances in Artificial Intelligence (AI) and Machine Learning (ML), as shown in Fig. 1. Currently, government incentives and technology advances have led to the large-scale deployment of sensing devices like smart meters and Phasor Measurement Units (PMUs) or micro-PMUs [4]. In power grids, high-resolution measurements can capture the system dynamics, including the event process. For example, the resolution for the PMUs can vary from 30 to 120 samples per second, which is accurate enough to reveal the event dynamics [5]. Therefore, PMU-based or micro PMU-based situational awareness becomes popular for modern power systems.

For research on power situational awareness, there are two main categories [6]: the signal-analysis approach and the ML approach. *Signal analysis* employs signal processing techniques to understand the data and identify events, such as window-based thresholds [7], wavelet transformations [8], and Swing Door Trending (SDT) [6]. These methods can accurately obtain the event time along the multivariate time-series, but lack the ability to identify the event types and locations. *Machine learning* methods extract underlying features of PMU data to indicate event time, types, and locations. Specifically, ML supervised learning methods can simultaneously build feature variables and link them to the event type or location labels. For example, ML methods are successfully applied to power system event identification, including Decision Tree (DT) [9], Support Vector Machine (SVM) [10], Artificial Neural Network (ANN) [11,12], Long Short-Term Memory (LSTM) units [13], and hybrid machine learning [14]. However, these methods mainly focus on fast event identification after the appearance of the event, but lack the ability to predict the event before it happens.

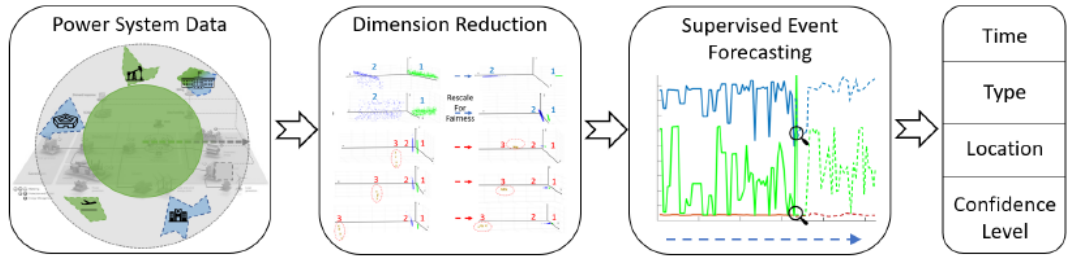

Fig. 1. Power Grid Analysis with Machine/Deep Learning

### 1.1 Power Event Forecasting Questions

An *event forecasting* ability is of greater importance to the public and government energy grids since these grids require high amounts of power supply and prefer that no power failure occurs at any time [15]. Therefore, it's important to understand the possibility of event forecasting in public grids. On the other hand, as illustrated in [16], the AI-based method should deliver certain accountability and explainability to societal, governmental, and public sector applications. Thus, three fundamental questions to boost the AI/ML applications in the energy sector include: (1) can a system identify the pre-event patterns to forecast a fault in the electrical grids? (2) Can a system guarantee the speed of the forecasting so that it can be realistically used? And (3) are there measures for the confidence of the forecasting?

To answer the questions of identification, forecasting, and confidence, there is a need to fully understand the phasor measurements with physics. Especially, for question (1) of *event identification*, the overloading of a line current can imply that there may be a line trip. Certain types of measurement disturbances may indicate the oscillation of the line sag, which may further cause the line to a ground fault. Thus, there is a need to properly mine the underlying patterns to forecast an event.

For question (2) of *forecasting speed*, it is observed that the PMU or micro-PMU measurements have a large data volume due to the high resolution, which may prevent the rapid forecasting algorithm. On the other hand, many PMUs may have similar event patterns to a specific event, e.g., the frequency response to a generator trip. Thus, it's essential to reduce the PMU dimensionality and extract the important features in a low-dimensional feature space.

For question (3) of *confidence assessment*, there are many approaches within ML and sensor data fusion that support confidence through information theoretic, probabilistic, and belief assessment. One method is a fusion of classifiers towards increasing the confidence reliability in the information knowledge that provide a well-defined utility measure to support decision making. Hence, for government and private sector deployment of AI/ML systems, there is a need to not only build a model for prediction, but also provide a measure of uncertainty in the decision. The measure of uncertainty should be based both on the data-driven assessment, the AI/ML method selected, as well as the physical modeling and constraints from which the data was collected. Thus, physics–informed, -enhanced, and -consistent ML is desired.

### 1.2 Power Event Modeling with Physics-Enhanced ML

Dynamic Data Driven Applications Systems (DDDAS) seeks to leverage real-time measurements with that of physical simulations such as that for power grids [17, 18]. The physics-based simulations provide a safety measure for public AI and reduce reliance on incomplete and faulty measurements for secure power grids [19]. For future power grid safety, DDDAS supports self-healing [20] and operational planning [21]. Coupled with signal-based physical methods are those that employ data-driven ML.

One aspect of the benefits of ML is data dimensionality reduction, where Chen *et al.* [22] identify PMU measurements with large variances via principal component analysis (PCA). Li *et al.* [23] builds an event-type dictionary with a subspace spanned from the principal components. Nguyen *et al.* [9] employs hierarchical clustering to group the dynamic behavior of generators. All of the above mentioned methods provide dimensional reduction, but fail to provide an accurate guarantee of following event prediction. Namely, the methods typically focus on the optimization of preserving enough information after dimensionality reduction, but lack the effect on targeting the event forecasting. Further, due to the heterogeneous dimensions, including the temporal, spatial, and measurement dimensions, there is a need to carefully study how to efficiently extract the useful information.

This paper proposes a general framework for fast event forecasting, including time detection, type differentiation, and failure localization as *power system event forecasting (PEF)*. Firstly, to tackle the multiple dimensions, a PCA-based technique efficiently captures the inner-correlations among different dimensions. Specifically, the PCA method utilizes a container to store the phasor measurements, where each measurement value is assigned a specific time slot, on a specific node, and indicates a specific measurement type (i.e., voltage, current, frequency, and power, etc.). Secondly, PEF leverages supervised learning-based approaches to map the low-dimensional data to the corresponding event labels. With many supervised learning (SL) techniques, PEF doesn't need to distinguish which one of the best models the power data. Instead, with data fusion, multiple learning methods are aggregated with a decision mechanism [24]. Finally, based on data fusion (e.g., voting [25]), an entropy-based measure characterizes the confidence level. Comprehensive experiments demonstrate the high accuracy and efficiency of PEF methods as compared to other state-of-the-art ML methods.

The rest of the paper is organized as follows. Section 2 provides a motivation and Section 3 the power grid definition. The proposed ML framework is explained in Section 4. Section 5 demonstrates the experiment setting and the representative results of the methods. Section 6 provides a discussion and Section 7 offers conclusions and future work.

## 2. Motivation

In the public, private, and industrial development of AI/ML systems, there is a need to develop the standards, implementation, and training procedures to deploy the tools and techniques. Past efforts have motivated the foundation for a standards approach for AI/ML deployment of sensor systems [26, 27].

For example, combining data from different sources, models, and sensor readings constitutes information fusion. The standard processing approach comes from the Data Fusion Information Group (DFIG) model which defines six levels of information fusion [28,29,30]. For example, Level 1 information fusion includes *object assessment* which includes determining the time and location [31] correlations as well as feature association [32,33]. Over the many debates of trends [34], how to score advances in data fusion and ML are a subject of debate and continual research.

Recent issues of moving from ML to DL motivate discussions to revisit the notion of Level 2 information fusion of situation assessment (SA) and Level 3 threat assessment. For SA, there are a variety of interesting developments that need further clarification for the deployment of AI/ML systems in a DFIG context. For example, advances in large data processing are subject to connectivity [35], adversarial [36], edge [37], and operational [38] constraints. Examples of such data fusion network challenges include network security [39], data privacy [40], and control access [41].

Hence, recent advancements in ML and DL have ushered in new opportunities for network computing [42], complex situation analysis [43], and long-term reasoning and planning for Level 4 information fusion. For example, Roy *et al.* [44] looked at ML/DL approaches for the signal transmission of radar to uncover and collect signals. Munir *et al.* [45] addressed the architectural needs for data fusion methods in Fog servers.

Another approach is Level 5 information fusion of *human-machine teaming*, which supports the adoption of systems. Recently, Braines, *et al.* [46,47,48] focused on how ML/DL sensor data fusion systems are deployable.

Together, research in AI (i.e., ML, DL) connections with data fusion and operator involvement requires coordination with the domain application. One example is that of the power and energy grid analysis deployment of AI. Hence, the public and private interest in the power and energy grid networks provides a motivating example of *networked AI*.

## 3. Power Grid Problem Definition

Fig. 1 shows a general idea of the proposed PEF framework. PEF analyzes the phasor measurements of the public grid to identify (1) when the event will occur, (2) what type the event is, and (3) the location of the event. To achieve these goals, PEF uses data dimensionality reduction and event forecasting processing.

As shown in Fig. 2, forecasting an event requires a series of data and the corresponding event label. For example, if there is an event in the historical data, the pre-event data is stored in a matrix $X = \Re^{A \times B}$, representing the PMU measurements in $A$ time slots with a number of $B$ different measurement types. The storage follows a moving-window-based manner, as shown in Fig. 2. Namely, a moving window with length $A$ is introduced to segment the PMU measurements for the storage. Therefore, the PEF model will assess each moving window and provide the time, type, and location of the upcoming event. Thus, a label vector $y$ contains all the information for power grid analysis. Formally, the following problem definition is:

- *Problem*: Event Forecasting via Machine Learning
- *Given*: PMU data of $X$ and the label vector $y$.
- *Find*: The mapping rule of $f: X \rightarrow y$.

Note that the mapping rule in the PEF setting is an abstract function that includes the dimensionality reduction and the mapping of data to labels.

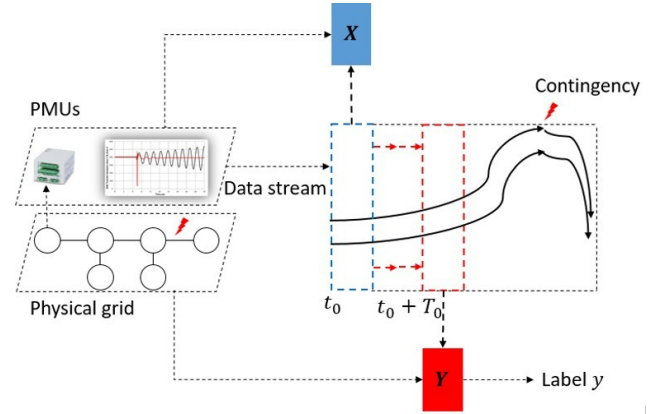

Fig. 2. The moving window for data storage.

## 4. Machine Learning

As shown in Fig. 1, PEF first reduces the dimensionality of the phasor data to achieve real-time forecasting. Thus, this section highlights the popular Principal Component Analysis (PCA) method for dimensionality reduction.

### A. Dimensionality Reduction

There are many ways to conduct dimensionality reduction. The general idea is to reduce the redundancy of the data and project the high-dimensional data to a low-dimensional space, while preserving as much information of the raw data as possible. For extracting the most important features, the popular PCA is utilized to find the major-variance directions.

Without loss of generality, assume $X$ is centralized, leading to the direct application of the singular value decomposition (SVD) for PCA:

$$X = USV^T, \qquad (1)$$
$$Y = XV$$

where columns of $V$ represent principal components and $U$ and $S$ are designed to project the original matrix into the PCA-based space to obtain data matrix $Y$.

The issue with the PCA method (1) is that the feature columns in the $X$ matrix are heterogeneous for the space and the measurement-type dimensions. Thus, treating them as homogeneous features for linear summation may not help capture the underlying complex correlations. Therefore, new tools are needed to maintain the physical consistency of the data and mine the complex non-linear correlations. Possible improvements include proposing to utilize tensor decomposition for such an efficient feature extraction.

Table 1 lists the advantages and disadvantages of machine learning (ML) methods [49].

Table 1: Machine Learning Methods

| Method | Advantage | Disadvantage |
|---|---|---|
| Decision Tree | Layered Tree Structure, White Box for Explanation | Inflexible Representation, Easy to Face Overfitting |
| K-Nearest Neighbors | Simple and Nonparametric | Long Computational Time, Sensitive to Local Data Structure |
| Naive Bayes | Theoretical Support from Bayes' Theorem | Strong Assumption on Conditional Independence |
| Logistic Regression | Fewer Assumptions and Direct Learning | Complex Model, Hard to Explain |
| Support Vector Machine | Maximum Margin, Easy to Calculate with Kernel Trick | Need to Choose a Good Kernel |
| Principal Component Analysis | Discovers patterns in data, provides feature dimension reduction | Requires standardization, Knowledge loss, less interpretable |

*B. Event Forecasting Using Supervised Learning Models*

After data preprocessing, supervised learning (SL) can be used to study the relationship between PMU measurements and related event time, types, and locations. For example, a transformer near the university may fail due to the extreme weather, causing the power outage to the university. Thus, there is a need to identify the location of the transformer quickly using SL methods. To better understand the example, consider using different learning methods as candidates for deciding the upcoming events given the current data. Fig. 3 shows the process of the proposed hybrid ML methods with data fusion. From the left side, the hybrid learning method imports event log files, including historical PMU measurements and labels, into different ML models. Each ML model trains its classifier separately. To avoid biases from different learning models, the process in Fig. 3 not only fuses the results of different models (e.g., classifier fusion) for event identification but also provides an entropy-based index to measure the confidence of the voted label. The Index $E$ is:

$$E = 1 + \frac{1}{N}\sum_{n=1}^{N}\sum_{k=1}^{K}\frac{p_{(n,k)}\text{Log}(p_{(n,k)})}{\text{Log}(M)} \qquad (2)$$

where $M$ is the number of classifiers, $N$ is the total number of testing samples, $K$ is the number of data inputs to the fusion process (e.g., voting) and $p_{(n,k)}$ is the weighted value for label $k$ in the $n^{th}$ testing sample. For each test example, if different ML methods agree, the system will obtain an entropy of 0 with the index $E = 1$, giving high confidence. If the classifiers differ among the labels, the confidence entropy would be Log($M$). With normalization and subtraction in (2), the index is $E = 0$, showing that there is little confidence in the estimate. Since entropy is unitless, it needs to be translated into another scale (e.g., probability) for decision making such as a receiver operating characteristic (ROC) curve [50]. With the ROC, the decision can be made based on a threshold.

Data fusion (DF) has many approaches, each with strengths and weaknesses. For example, evidential reasoning handles conflicting and data uncertainty in multiple classifier outputs. Many of these DF approaches can be implemented to determine which method would work best in coordination with AI/ML systems. Table 2 lists different data fusion methods [51].

Table 2: Data Fusion Methods

| Method | Advantage | Disadvantage |
|---|---|---|
| Boolean | Well defined, expandable | Lose meaning, rules not address reasoning |
| Fuzzy Logic | Handle partial information, possibility theory | Ill-defined IF-THEN rules, large number of rules needed |
| Bayes Networks | Uncertainty modeling, propagation of information in network | Requires a priori knowledge, highly complex relationships |
| Markov Chains | Simple to define, scalable, handle uncertainty | Knowledge needed to define, prune, and transition states |
| Entropy | Unitless, predict uncertainty | Misinterpreted, results require probability translation |
| Evidential Reasoning | Handles conflicting, negative information | approximations for set, partial information, complex |

The challenge for the public and/or government selection and adoption of a specific approach would require testing and evaluation towards first verifying the methods and then validating in the domain of choice. A better approach would be to adopt certain standards, even if a baseline, so as to provide consistency. For example, entropy analysis is widely used and can be a standard measure incorporated into systems. Such an ontology of metrics and definitions would help in the standards analysis [52,53]. While additional methods and metrics can be added to the ontology, at least a common approach is required for adoption, certification, and deployment of AI systems towards meeting a common standard.

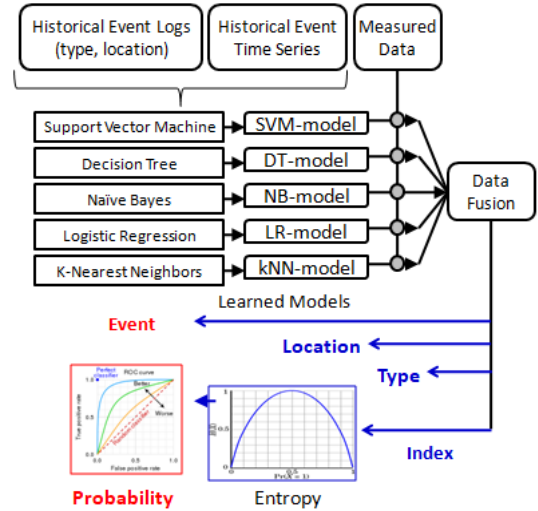

Fig. 3. Machine Learning and Data Fusion

Additional concerns for the adoption of AI analysis should include data privacy [54] and security [55]. For example, in the power grid analysis, security and integrity of the data should be maintained with the index when forecasting events.

# 5. Experiments

The performance of the proposed PEF framework and other benchmarks are evaluated with five different power systems, where a 20% penetration of sensors (i.e., 20% of nodes are equipped with sensors) is presumed due to the expensive sensor cost. In addition, we test the PEF model effectiveness when sensor penetration varies from 5% to 30%. We show the PEF ML-model comparisons as baseline methods towards future experiments using DL methods. All experiments are conducted under a 64-bit machine with Intel(R) Core(TM), i7-6700HQ, 2.6 GHz CPU, and 16 GB memory.

*A. Experiment Setup*

In the study, we consider a common line failure in powergrids. When the transmission line is overloaded for a length of time, the line may be tripped, decreasing the system stability which may cause a blackout. Thus, forecasting this contingency is essential for social benefits.

*Data acquirement*: in power systems, the system-state data can be directly measured or simulated with a large set of differential equations, where the customer power consumption is the input variable, grid parameters (e.g., grid topology and the line impedance) are equation parameters, and nodal voltages are the demanded state-variable solutions. To solve the equations, the power community develops many high-level simulators. Notably, we employ a business-level simulator, Positive Sequence Load Flow (PSLF) [56] from General Electric (GE) for data acquirement.

*Real-world customer power consumption*: the experiment uses realistic data [57] from PJM Interconnection LLC. The dataset contains the 2017 year's power consumption data in the PJM Regional Transmission Organization (RTO) region. Fig. 4 visualizes the daily power curves at one bus in 2017. Fig. 4 shows high power consumption from 14:00 to 20:00, and for some days, the power consumption is larger than 9000 kWh, which is the potential area for the overloading of the line connected to that bus. Further, different buses have different power consumption patterns, and they work together with complex grid parameters to determine the system state.

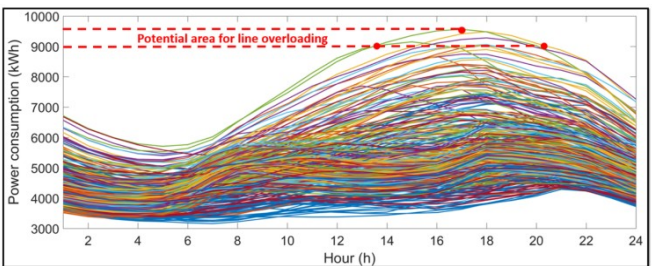

Fig. 4. Visualization of the PJM power consumption data.

*Grid parameters and simulation setting*: we introduce five standard power systems, namely IEEE 30-, 85-, 300-,1888- bus systems [58], and the Illinois 200-bus system [59]. The topology of the Illinois 200-bus system is shown in Fig. 5, which shows the network connections of the power distribution.

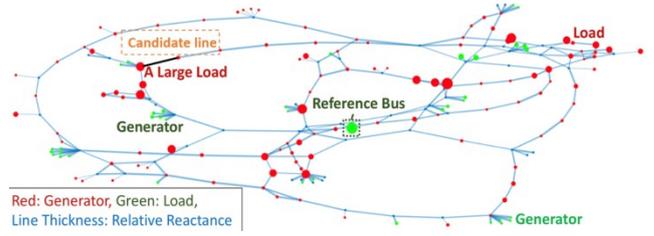

Fig. 5. Topology of Illinois 200-bus system.

For every system, assume there is a 20% penetration of sensors and the system randomly selects the sensor-equipped nodes for five times to obtain the average result for a fair evaluation. For each sensor, we assume there are four measurement types for each sensor (i.e., voltage magnitude, voltage angle, active power injection, and reactive power injection). Then, we employ a moving window to check the time series, each of which contains 166 time slots. In each grid, four candidate lines are considered for overloading and, potentially, line trips. For example, in the Illinois 200-bus system, line 15-16, 29-30, 123-125, and 149-152 are connected to large-load buses with a high probability for overloading and line trips. Once they are overloaded, the simulator will cut the line. Then, there are formed a one- or multi- (i.e., multiple lines are cut simultaneously) hot label vector $\mathcal{Y} \in \{0,1\}^{5 \times 1}$ to represent the later contingency status, where a 1 for the first element represents the normal status but for the other elements, it represents the corresponding line-trip contingency.

*Methods*: due to the ultra-high dimensionality of the dataset, the PCA is employed to reduce the dimensions. Different number of PCAs, ranging from 50 to 2000, are selected and input to advanced multi-class learning methods, namely: "One-vs-the-rest" Support Vector Machine (OVRSVM), Multi-label Logistic Regression (MLLR), and a popular deep learning (DL) method Convolutional Neural Network (CNN) used for image processing [60], as shown in Fig. 6.

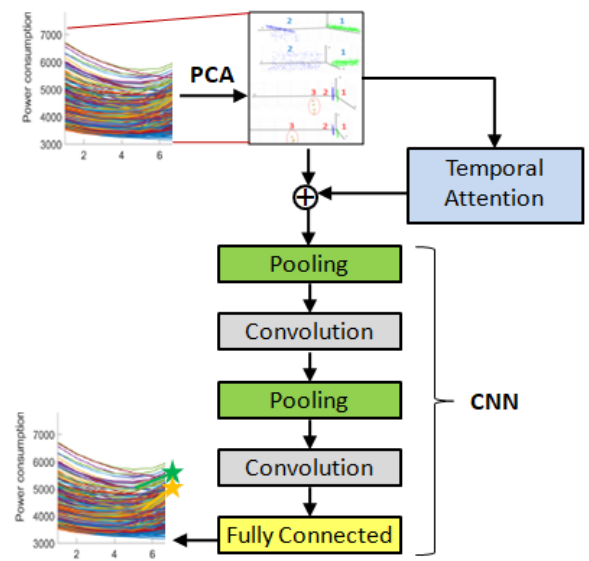

Fig. 6. PCA+CNN Architecture

*Performance Evaluation*: for each method, we conduct the 3 fold cross-validation. To evaluate the performance, the Mean Zero-one Error (MZE) is utilized:

$$MZE = \frac{1}{N} \sum_{n=1}^{N} [\![y_i^* \neq y_i]\!] \quad (4)$$

where $[\![\bullet]\!]$ represents the operation to count when the inside term is true, $y_i^*$ is the predicted label, and $y_i$ is the true label. For each method, PFE calculates the MZE for each validation set in the 3-fold validation and averages them as their final output.

## B. Performance

There is a need for classification accuracy with stable, high performance. We test five different smart grids and Table 3 presents the MZE value and the number of parameters for each model. In general, for classification accuracy, the PEF framework verifies the performance under all scenarios, where ML uses few parameters and achieves better results than the PCA+CNN.

Table 3: Results: N Bus-System (BS) performance for each method (MZE Value/No. of Parameters)

| N/Method PCA+ | OVRSVM | MLLR | CNN |
|---|---|---|---|
| 30Bus | **0.105/151** | 0.124 / 151 | 0.350 / 1685 |
| 85 Bus | 0.420 / 451 | **0.345 / 151** | 0.540 / 1594 |
| 200 Bus | **0.331 / 351** | 0.423 / 301 | 0.402 / 2333 |
| 300 Bus | 0.261 / 401 | **0.260 / 451** | 0.520 / 1354 |
| 1888 Bus | 0.350 / 1001 | **0.303 / 1001** | 0.499 / 4274 |

The average improvement of the PEF framework over the best benchmark methods is 10%. We notice that for different grids, the best benchmark method varies while the framework is robust with the best classification performance. We discover that the number of parameters of the framework is usually less than normal neural networks but larger than weak learners, which shows moderate model complexity to avoid overfitting. In summary, the PEF framework achieves good performance in classification accuracy and robustness to different networks with moderate model complexity.

*Robust performance under different sensor penetrations*:
To further test the model robustness, we consider different sensor penetrations, ranging from 0% to 30% for the IEEE 300-bus system. Fig. 7 illustrates the results. Our PEF framework presents the lowest MZE value and as the penetration increases, it keeps decreasing from 10% to 20%, with only a slight increase from 20% to 30%, i.e., 30% is a corner point for sensor penetration. For the baseline methods, they have higher classification errors and their corner points range from 15% to 25%.

## 6. Discussion

Much of the discussion of the deployment of AI is whether the AI successes are ready for adoption in the public sector for decision making. Since the AI methods typically focus on data-driven approaches, there is a need to consider the domain of application, improvement and/or coordination with physical knowledge, and data limitations. For the domain, the paper explores the power grid. Much physical knowledge is known from power grid modeling, which could be used as a digital twin in the DDDAS framework. The challenge is when to use the current data and physical models such as for estimation and when to use for forecasting. If the systems employing AI are autonomous (i.e., without human intervention) arbitrating the use of AI [61] is still a subject that requires further research towards physics-based consistency [62].

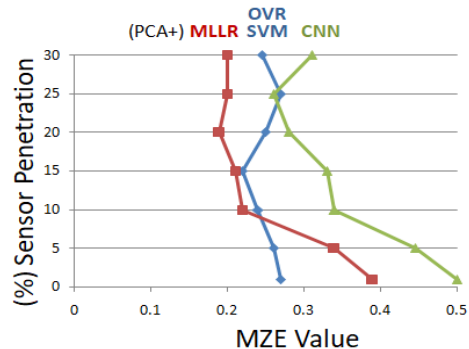

Fig. 7. Sensor Penetration versus MZE value.

## 7. Conclusions

This paper focuses on event forecasting of public power grids and discovers a classification mapping from pre-contingency measurements (i.e., precursors) to the later contingency status. However, there is no systematic design to tackle the high dimensionality and forecast an accurate label. To handle these challenges, the PEF framework is proposed with (1) dimensionality reduction to preprocess the data and (2) an efficient data fusion ML model to map the processed data to labels. In the experiment, we introduce the power systems data and conduct comprehensive tests. The final result shows that the proposed PEF framework provides improvement. This opens the door for public grid event forecasting for real-time operation, but further work is needed with enhanced deep learning and data fusion methodologies.

### Acknowledgments

The views and conclusions contained herein are those of the authors and should not be interpreted as necessarily representing the official policies or endorsements, either expressed or implied, by author affiliations.